# Geometrical Complexity of Data Approximators

Evgeny M. Mirkes[1], Andrei Zinovyev[2,3,4], Alexander N. Gorban[1]

[1]Department of Mathematics, University of Leicester, UK,
{em322,ag153}@le.ac.uk
[2]Institut Curie, rue d'Ulm 26, Paris, 75005, France
[3]INSERM U900, Paris, France
[4]Mines ParisTech, Fontainebleau, France
andrei.zinovyev@curie.fr

**Abstract.** There are many methods developed to approximate a cloud of vectors embedded in high-dimensional space by simpler objects: starting from principal points and linear manifolds to self-organizing maps, neural gas, elastic maps, various types of principal curves and principal trees, and so on. For each type of approximators the measure of the approximator complexity was developed too. These measures are necessary to find the balance between accuracy and complexity and to define the optimal approximations of a given type. We propose a measure of complexity (geometrical complexity) which is applicable to approximators of several types and which allows comparing data approximations of different types.

**Keywords**. Data analysis, Approximation algorithms, Data structures, Data complexity, Model selection

## 1 Introduction

### 1.1 Complexity of Data as Complexity of Approximator

It would be useful to measure the complexity of datasets which can be represented as a set of vectors in a potentially high-dimensional space. In this study, we analyse a measure of data complexity based on the analysis of the geometry of the data approximators. Therefore, we call it the geometrical measure of complexity.

There are many ways to define the complexity of data distribution. For example, Akaike information criterion (AIC) can be used to select models of data of minimal complexity, using information theory [1], or Structural risk minimization principle is applied in some areas [2].

Recently we proposed to measure data complexity through complexity of data approximators and developed three measures of data complexity: geometrical complexity, structural complexity and construction complexity [3-5].

Actually, each data approximation method is equipped with a measure of the approximator complexity. For example, in the classical data approximation methods, the number of centroids in k-means clustering, the number of principal components in Principal Component Analysis, and curvature or length of the principal curve serve as measures of complexity. A close to optimal approximator must be able to catch the hypothetical intrinsic shape of the data distribution without trying to approximate the data's "noise" (though "one man's noise is another man's signal" [6]).

## 1.2 Complexity of Data Approximators

Measuring complexity of data approximators is important because of the following principle, which is an application of the famous Occam's razor: between two approximators of the same accuracy one should prefer the one with smaller complexity [7]. Thus, complexity measure becomes a computational tool for selecting the best approximator type and structure, a tool for model selection. The model selection problem is the classical problem of mathematical modelling and statistics. Its solution is always based on the complexity/accuracy trade-off [8]. Model selection should be based not solely on goodness-of-fit, but must also consider model complexity [9]. The positive effect of model complexity on estimation error for new data points from the same data domain can be directly demonstrated by cross-validation computational experiments in some settings [9].

Selecting a model of data with the least complexity can be even more crucial for *model generalizability* on new data domains not accessible through available sampling [10]. This model ability cannot be directly demonstrated by cross-validation. Using complexity criteria here gives us hope that the simpler model will better fit outside its definition domain, sometimes at the expense of worse accuracy for available data (accuracy/generalizability trade-off). However, this remains a strong hypothesis in statistics. In other fields of science (for example, theoretical physics) the internal simplicity ("beauty") of a model or a theory often guides theoretical constructions with many examples of success.

However, measuring and even defining complexity is a difficult task. Many methodological studies converge on that the notion of *complexity* should be distinguished from the notions of *size*, *order* and *variety*, and that any measure of complexity is dependent on the language of object's representation [11]. The most common attributes used to compare alternative models are the level of detail and the model complexity although these terms are used in many different ways [12]. There may be proposed many complexity criteria. Each of them represents an aspect of complexity and together they can be considered as a representation of the "objective complexity". At least, there is no other "objective complexity" besides a combination of various technical complexity measures.

Coming back to the problem of data approximation, we can distinguish two important characteristics of any approximator: 1) its "internal" structure, and 2) its "flexibility", i.e. its ability to be tuned in order to fit the data. For example, a polynomial approximator is characterized by its maximal degree (internal structure), and the constraints on the parameter values (flexibility). Therefore, talking about approximator's complexity, we can distinguish the complexity of the approximator's structure and the complexity of its configuration in the data space.

A large class of data approximators can be represented in the form of graphs, embedded into a data space [3], [13]. Here one can talk about the complexity of the graph structure itself and the complexity of the mapping of the graph's nodes into the data space. For example, a one-dimensional grid with k nodes can represent both linear and curvilinear approximations of data, with linear mapping being evidently less complex.

In this paper, we compare approximators represented by graphs of two different types: one-dimensional grids and trees. From what was said above, we have two options to compare them in terms of complexity: the complexity of their structures and the complexity of their mappings. Comparing structures of graph-based approximators is not always meaningful: they can be generated using different types of generating grammars (description languages). Here, the grammar generating trees has more types of elementary transformations (more variety) than the one generating one-dimensional grids.

In this study we focus on comparing approximators for their mapping complexity rather than comparing their structures. In this paper we present and test a concrete recipe for this, *the geometrical complexity*.

## 1.3     Three Measures of Approximators' Complexity

Three measures of the approximator's complexity were introduced in [3-5]: (1) Geometrical measure, (2) Structural measure, (3) Construction measure.

*Structural and construction complexity measures* introduced below estimate the complexity of the "internal" structure of the approximating graph, while *geometrical complexity* is an analogue of non-linearity for the graph-based approximators.

**Geometrical Complexity.** The geometrical measure of complexity estimates the deviation of the approximating object from some *"idealized" configuration*. The simplest such ideal configuration is linear: in this case the nodes of the approximator are located on some linear surface. Deviation from the linear configuration would mean some degree of non-linearity.

However, the notion of non-linearity is applicable only to relatively simple situations. For example, in the case of branching data distributions, non-linearity is not applicable as a good measure of geometrical complexity. In [13] it was suggested that a good generalization of linearity as "idealized" configuration can be the notion of harmonicity. An embedding of a graph into linear space is called *harmonic* if, in each star of the graph, the position of the central node of the star coincides with the mean of its leaf vectors [13]. For many applications it is convenient to consider not all the stars but a set of selected stars and to introduce *pluriharmonic* embedding [5,13]. For pluriharmonic embedding, the position of the central node of each selected star coincides with the mean of its leaf vectors.

In our estimations of the geometrical complexity of trees we will use the deviation from a harmonic embedment as analogue and generalization of the non-linearity.

In [4] geometrical complexity was used to optimize the balance between accuracy and complexity (see Fig. 1).

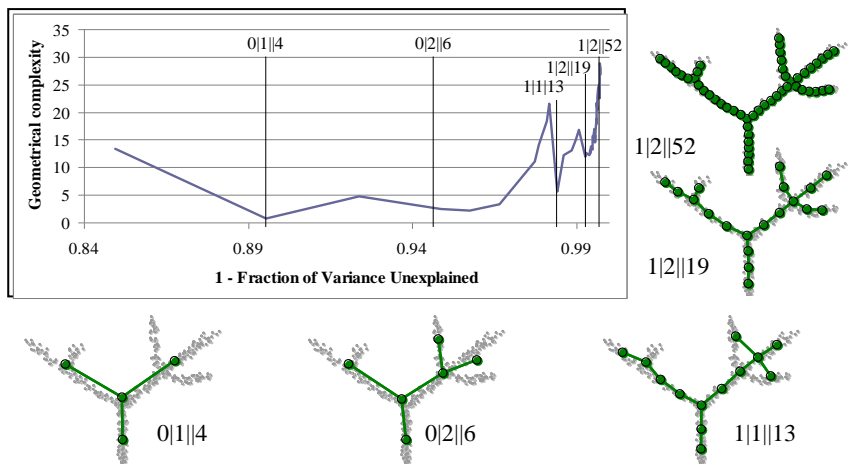

**Fig. 1.** The "accuracy–complexity" graph for a tree-like 2D data distribution. Several scales of the approximator's complexity are shown. Two of them, corresponding to the structural complexity barcodes 0|1||4 (1 3-star and 4 nodes) and 1|2||19 (1 4-star, 2 3-star and 19 nodes) are optimal and approximate the distribution structure at a certain "depth".

**Structural Complexity.** The structural complexity defines how complex is an approximator in terms of its structural elements (number of nodes, edges, stars of various degrees). In general, this index should be a non-decreasing function of these numbers. Contribution of some of the elements (for example, nodes and edges) might be not interesting for measuring the structural complexity and, hence, have zero weight (not present) in the resulting quantitative measure.

**Construction Complexity.** We derive our approximators by the systematic application of the graph grammar operations, in the way which is the most optimal in terms of the objective function. The construction complexity of an approximator can be defined as a minimum number of elementary graph transformations which were needed to produce it, using given grammar. This measure can be similar to the structural complexity in some implementations but is not equivalent to it. There is some similarity to the idea of the Kolmogorov complexity (see, for example, [14]) but the Kolmogorov complexity is defined on the base of the necessary length of the program and the construction complexity is the number of operation actually used in construction.

In this paper we show that the geometrical complexity can be used to compare approximators of different types. In particular, we compare Growing Self-Organising Maps and Growing Principal Trees in terms of this measure.

As for the other two measures, structural and construction complexities, it happens that for the two types of graph-based approximators considered in this paper, we can only compare them in terms of the number of nodes in the graph (see Table 1).

## 2 Materials and Methods

To check the ability of geometrical complexity (GC) to compare two different approximators, we designed several benchmarks datasets (see Fig. 2), which we approximated using two data approximation methods: Growing Self-Organizing Map (GSOM) [15] and Principal Tree (PT) [13].

First, for both methods we used the fraction of variance unexplained (FVU) as the measure of the approximation accuracy. In all tests we stopped the approximator's growth when the approximation accuracy was less or equal to a specified threshold value, which was the same for both GSOM and PT. For "*thin*" patterns the threshold value of FVU was equal to 0.1%, for "*scattered*" patterns the threshold value of FVU was equal to 0.2% and for "*scattered and noised*" patterns threshold value of FVU was equal to 1%.

The same parameters of approximation methods were used for all benchmarks.

For comparison of the constructed approximators we used three criteria: 1) number of nodes, 2) length of the constructed approximating graph, which, by definition, is the sum of all graph edge lengths, and 3) geometrical complexity.

### 2.1 Fraction of Variance Unexplained

The Fraction of Variance Unexplained (FVU) is the dimensionless least square evaluation of the approximation error. It is defined as the ratio between the sum of squared distances from data to the approximating line and the sum of squared distances from data to the mean point.

The distance from a point $x_i$ to a line is the distance $p_i$ between $x_i$ and its orthogonal projection onto the line. The FVU of a line, used as an approximator, is simply $\text{FVU} = \sum_{i=1}^{n} p_i^2 \bigg/ \sum_{i=1}^{n} \|x_i - \bar{x}\|^2$, where $\bar{x}$ is the mean point $\bar{x} = (1/n) \sum_{i=1}^{n} x_i$. For a polygonal line (an ordered set of nodes $\{y_i\}$ $(i = 1, 2, \cdots, k)$, connected by line fragments) FVU is defined in the following way. For a given set of nodes $\{y_i\}$ $(i = 1, 2, \cdots, k)$ we compute the distance from each data point $x_i$ to the polygonal line specified by a sequence of points $\{y_1, y_2, \cdots, y_k\}$. For this purpose, we calculate all the distances from $x_i$ to each segment $[y_s, y_{s+1}], s \in [1..k-1]$ and find the minimal distance $d(x_i)$. In this case we define $\text{FVU} = \sum_{i=1}^{n} d^2(x_i) \bigg/ \sum_{i=1}^{n} \|x_i - \bar{x}\|^2$.

### 2.2 Growing Self-Organizing Maps

Growing self-organizing maps is a well-known method to approximate sets of vectors embedded in high dimensional space by self-organizing maps with ability to increase the number of nodes [15]. For this study we use the liner version of GSOM. A de-tailed description of the algorithm used can be found in [16]: below we provide a brief description.

1. Initiate SOM by two nodes, connected by an edge, which is a linear fragment of the first principal component, centred at the mean point.
2. Optimize the node positions by using the Batch SOM learning algorithm [17].
3. Compute the current FVU (CFVU).
4. If CFVU is less than or equal to the specified threshold value then stop. If not then
5. Glue to the first node of the polygonal line a new edge, of the same length and direction as the first edge of the polygonal line, and calculate FVU for the new polygonal line (BFVU).
6. Extend the last node of the polygonal line with a new edge, of the same length and direction as the last edge of the polygonal line, and calculate FVU for new polygonal line (EFVU).
7. If BFVU<EFVU and BFVU<CFVU then glue to the first node of the polygonal line a new edge, of the same length and direction as the first edge of the polygonal line, and go to step 2.
8. If EFVU<BFVU and EFVU<CFVU then extend the last node of the polygonal line with a new edge, of the same length and direction as the last edge of the polygonal line, and go to step 2.

For each GSOM in this study we used the linear BSpline neighbourhood function with neighbourhood radius 3.

### 2.3 Principal Tree

Principal graphs form a rich family of tools for nonlinear dimensionality reduction for data of complex topology [18]. The method of topological grammars for construction of principal graphs and cubic complexes was proposed in [13]. The detailed description of principal tree (PT) construction algorithm can be found also in [4], [16]. The parameters which were used in this work are described below.

We used the graph growing grammar with two rules: "add new node to an existing node" and "bisect an existing edge". We used the graph shrinking grammar with two rules: "remove a leaf", "remove an edge, and glue its nodes". The grammar sequence "growing, growing, growing, shrink" was used for constructing the principal tree. In this study we used the elastic moduli 0.01 for stretching and 0.001 for bending.

1. Initiate PT by two nodes, connected by an edge, which is a linear fragment of the first principal component, centred at the mean point. Set stage = 1.
2. Optimize the energy functional of the graph
3. Calculate the current FVU. If FVU is less than or equal to the specified threshold value then stop. If not then
4. If stage < 4 then try all possible applications of growing grammar. Select the most optimal in terms of energy decrease operation. Apply the selected graph grammar operation. Add 1 to stage. Go to step 2.

5. If stage = 4 then try each possible application of shrinking grammar and save the best operation. Apply saved operation. Put stage = 1. Go to step 2.

## 2.4 Geometrical Complexity Used in the Study

A k-star is defined as a fragment of a graph, which is a node $v_0$ with all connected to it neighbour nodes $v_i : i = 1, \cdots, k$ (leaves). Let us denote by $v_i$ both the tree (polygonal line) node and the embedment (position) of the corresponding vector in the data space. For each embedment of a k-star into the data space, the measure of non-harmonicity is the squared deviation of the central node $v_0$ from the mean point of the leaf positions $\text{GC}(v_0) = \left\| v_0 - (1/k) \sum_{i=1}^{k} v_i \right\|^2$. This formula is analogous to the bending energy of a plate which is an integral of the squared first derivates of the deviation from a surface (k>2) or from a straight line (k=2). If we use an analogy with the method of finite elements then we have to multiply the sum of the bending energies in all k-stars by the squared number of nodes. Let us agree that for a 1-star the measure of non-harmonicity is equal to zero. Then the geometrical complexity of a tree (polygonal line) is $\text{GC} = n^2 \sum_{i=1}^{n} \text{GC}(v_i)$.

## 2.5 Available Implementations

A user-friendly graphical interface (Java-applet) for constructing principal trees and GSOM in 2D is available at [16].

## 3 Results and discussion

The resulting maps and trees for each benchmark are shown in Figs. 2 and 3. The numerical values of the complexity measures for them are listed in Table 1. Table 1 allows us to compare four characteristics of constructed approximators: the number of nodes N, FVU, their lengths, and GC. We can see that for the approximators with the same FVU three complexity measures may be significantly different. The difference between GC values may be huge (one order of magnitude) while the length and N do not differ significantly. Thus, one can conclude the GC measure can be more sensitive in distinguishing the least complex approximator.

    We can also compare two algorithms for approximation, GSOM and PT. For two patterns, "spiral thin" and "spiral scattered" they work similarly (the results are qualitatively the same, some measures are better for GSOM, and some for PT, without huge difference). The differences in the approximator lengths for the same patterns and given FVU are not big. The length of PT approximators is smaller in most cases.

The number of nodes for the same FVU is always smaller for PT, and GS of the PT approximators is almost always much smaller than for GSOM.

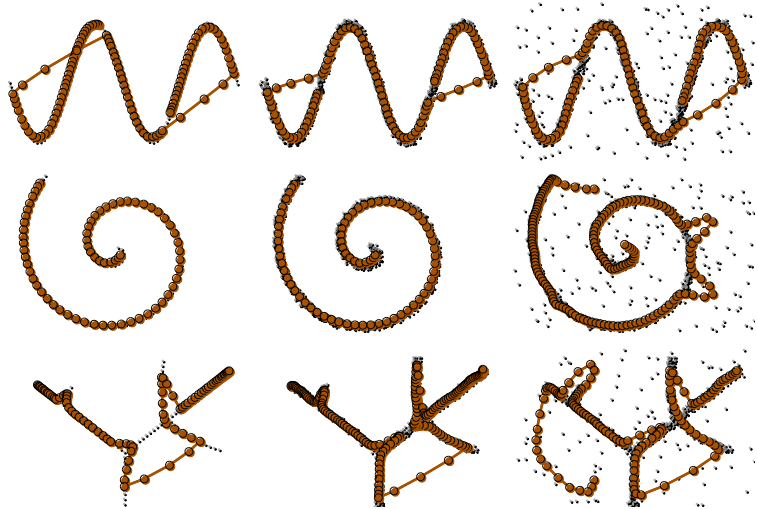

**Fig. 2.** Approximating polygonal lines constructed by GSOM.

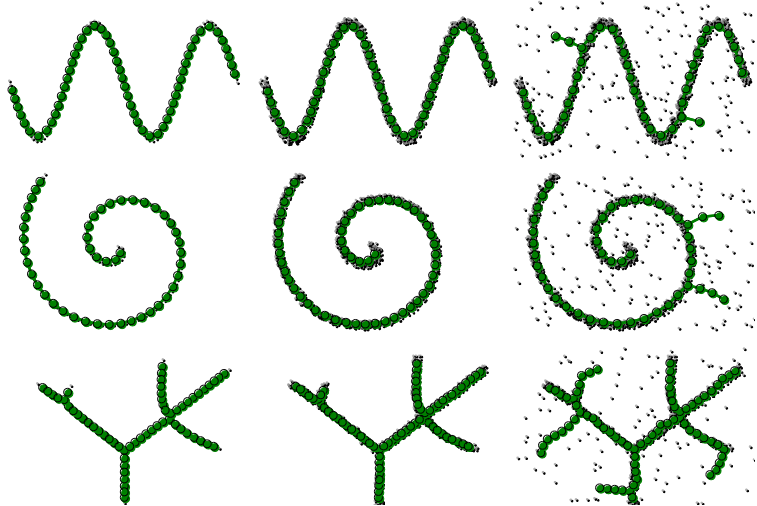

**Fig. 3.** Approximating trees constructed by the principal tree algorithm.

Table 1. Complexity measures for the constructed approximators. *N* is number of nodes which reflects the construction complexity measure in this case. In the column "Pattern" *S* means "scattered" pattern, *SN* means "scattered and noised" pattern. The winner value of the complexity measures for each pattern is highlighted in bold and underlined.

| Pattern | SOM | | | | PT | | | |
|---|---|---|---|---|---|---|---|---|
| | *N* | FVU | GC | Length | N | FVU | GC | Length |
| Sinus | 101 | 0.10% | 3 895.64 | 1 671.30 | **56** | 0.10% | **263.89** | **1 236.06** |
| Sinus S | 91 | 0.20% | 1 043.05 | 1 454.64 | **49** | 0.20% | **238.59** | **1 232.75** |
| Sinus SN | 90 | 0.98% | 939.10 | 1 551.04 | **41** | 1.00% | **325.26** | **1 357.12** |
| Spiral | 66 | 0.10% | **83.90** | **1 247.49** | **46** | 0.08% | 107.89 | 1 254.20 |
| Spiral S | 71 | 0.19% | **74.14** | 1 253.22 | **51** | 0.19% | 96.23 | **1 251.32** |
| Spiral SN | 158 | 0.99% | 3 164.63 | 1 680.79 | **46** | 0.99% | **166.84** | **1 442.07** |
| Tree | 82 | 1.05% | 1 951.71 | 1 101.49 | **59** | 0.09% | **25.17** | **973.18** |
| Tree S | 156 | 0.20% | 6 143.35 | 1 434.03 | **71** | 0.19% | **29.93** | **995.80** |
| Tree SN | 115 | 1.00% | 4 699.45 | 1 822.96 | **75** | 0.99% | **319.64** | **1 392.00** |

## 4   Conclusion

Accordingly to the general principles of model selection, among approximators having the same accuracy (FVU), the one with the smallest complexity should be preferred. Here we demonstrated that the geometrical complexity (GC) is applicable for comparing self-organising maps and principal tree approximators. Moreover, we show that the geometrical complexity investigated here is a more sensitive measure for model selection between these two types of graph-based approximators, compared to other measures such as the length and the number of nodes in the graph. Further investigations are required to demonstrate the utility of geometrical complexity for model selection among many different types of graph-based approximators.